\documentclass[letter]{spie}

\usepackage{graphicx} 
\usepackage{color}
\usepackage{enumitem}
\usepackage{makecell}
\usepackage{multirow}
\title{Interactive deep-learning based interface for brain tumour segmentation}
\author{Enora Giffard}
\author{Pierre Jannin}
\author{John S.H. Baxter}
\affil{Laboratoire Traitement du Signal et de l'Image (LTSI - INSERM UMR 1099), Université de Rennes, Rennes, France}

\begin{document}

\maketitle

\begin{abstract}
Brain tumour segmentation from MR images is an integral part of their treatment, either through surgical resection or radiotherapy. Both applications require a high degree of precision that can be difficult to ensure for all patients due to the high degree of variability in the size, shape, number, and location of these tumours. This leads to difficulties with their automatic segmentation via deep learning, which can display catastrophic segmentation errors when presented with images that differ, even subtly, in terms of the specific MR sequence used, which varies from vendor to vendor. This article presents a deep-learning-based interactive segmentation system that, while being fully automatic by default, has the capability of being easily corrected by human users, allowing them ensure sufficiently high brain tumour segmentation accuracy for all patients. This interface is evaluated both in terms of the accuracy of its automatic approach as well as its usability through a preliminary observational expert user study.
\end{abstract}

\section{Introduction}

Gliomas, being a common form of primary brain tumour (over 24\%) and the most common form of malignant brain tumour (over 80\%), represent a growing concern for the healthcare systems which must contend with their clinical management \cite{pellerino2022epidemiology}. Because of this, precise glioma segmentation in MRI is crucial as it forms the basis for tumour monitoring as well as surgical and radiotherapy planning. Segmentation also allows for more constructive conversations between clinicians and patients, using segmentation as the backbone for advanced computer graphic representations of the patient's anatomy which can be used to display tumour progression or to illustrate potential treatment options \cite{wang2022virtual}.

Manual segmentation is difficult and time-consuming, making it difficult for clinicians to perform during consultations or their otherwise already busy schedule. This has led to a heavy focus in the research community for fully automated segmentation algorithms, such as those based on recent deep learning paradigms, to alleviate these time concerns. Regardless of how well these automated algorithms perform, they are still susceptible to errors, especially large errors caused by sometimes imperceptible changes in the MRI caused by a variety of factors such as updated sequence design and motion artefacts \cite{hoebel2022know}. Thus, automatic segmentation can fail and result in the clinicians having to still perform manual segmentation, although this manual segmentation on particularly difficult cases can allow for a human-in-the-loop style of training an evolutive machine learning model in practice \cite{sailunaz2023brain}.

One potential avenue to alleviate this issue with automated segmentation is to design it in a way that incorporates interaction mechanisms, intuitive ways in which a clinical user can correct a segmentation algorithm at a particular scale corresponding to the level of error \cite{baxter2018semiotics}. If designed well, these interaction mechanisms can be directly incorporated into the functioning of the automatic algorithm itself, allowing for a singular framework to be both fully automatic by default and interactive when necessary \cite{baxter2022combining}. These interfaces would allow for a higher level of efficiency in practice than standard interactive segmentation paradigms (even those that use deep learning, such as that presented by Wei \textit{et al.}\cite{wei2023towards}) which do not have a high accuracy, fully automated default mechanism.

\subsubsection*{Contribution}
The purpose of this study is to extend our previous methods for interactive subcortical anatomy segmentation \cite{baxter2022combining,giffard2025interactive} to a more variable context, the segmentation of brain tumours in multi-sequence MR images. As per our previous work, this includes both the development of machine-learning based frameworks for addressing the problem of approximating the location and extent of the tumour quickly with another light-weight network for performing the fine-grained segmentation within that localised region. This is followed both by evaluation of the system's accuracy as an automated segmentation method as well as a user study to investigate how expert clinicians would use such a system in practice.

\section{Material and Methods}

\subsection{Data}

The data we use in this study are from the B.R.A.T.S. 2021 \cite{baid2021rsnaasnrmiccaibrats2021benchmark, Bakas2017TCGA, Menze2015BRATS} dataset. This data set contains images from 1251 patients, including T1 weighted images with and without gadolinium injection, T2 weighted images, and FLAIR sequences for each patient. Three labels are available in the ground truth, in addition to the background, enhancing the tumour, necrotic core, and peritumoral oedema. The images are padded to a constant size of 256x256x256 voxels. Due to the two-step nature of our framework and the need to draw a bounding box significantly smaller than the image around the tumour, we only use patient images in which the tumour is less than 96 voxels in any axis, which represents 743 patients. The full dataset is then split into a training dataset of 700 patients, a validation dataset of 8 patients and a test dataset of 35 patients.

\subsection{Pre-segmentation network}
Unlike our previous approaches to subcortical anatomy segmentation, brain tumours are widely known to have much more variable sizes (depending on tumour severity) and numbers (due to metastases), meaning that a simple centroid-based localisation framework cannot fully estimate the region-of-interest in the image containing the tumour. However, performing a fine-grained segmentation of the entire image would be very time-consuming which could limit use in practice. Therefore, we localise the tumours through the use of a \textit{pre-segmentation}, that is, a rough segmentation which only uses partial information in order to be quickly computable. Our pre-segmentation network is made up of a standard 3D U-Net \cite{ronneberger2015unet} which performs a binary segmentation of all tumours using the FLAIR and T2 images as input, which have been downsampled to an isotropic voxel resolution of 4 mm. This downsampling allows the image to be small enough to be quickly segmented in its entirety without patching. 

Once the pre-segmentation network has performed a coarse segmentation of the potentially multiple tumours, the segmentation map is thresholded at 0.99 and connected components analysis is performed on the result as to determine the number of tumours. Components with a volume of less than 100 voxels, which are likely artefacts, are deleted. If multiple components are present, their minimum bidirectional surface-surface distance is computed and, if it is less than 10 voxels, they are merged. If multiple tumours are still present, if some of them have a maximum intensity of over 0.99999 while others' maximum intensity is under 0.99999, the latter are deleted as they are likely artefacts. For each tumour, an axis-aligned bounding box is defined that contains the entirety of the segmented voxels and thus defines the region surrounding the tumour. This bounding box is then added to a user-visible list in the interface and a cropped copy of the image is made according to each bounding box, which are then passed to the fine-grained segmentation network.
In theory, if the segmentation drastically fails, it is likely to be at this level. Thus, we allow the user to add bounding boxes to the aforementioned list, modify those already within the list, and delete bounding boxes that do not refer to tumours but potentially image artefacts.

\subsection{Segmentation networks}
The fine-grained segmentation made use of a standard 3D U-Net \cite{ronneberger2015unet} with four resolution layers, slightly modified in order to take into account the variable size of the input image due to the aforementioned cropping. The lowest layer of this UNet made use of a fully-connected layer but, instead of being defined over all pixels from the previous layer, it instead used the average value in each channel. The output of said layer is then tiled to the input size; thus allowing information to flow across the entire image while also remaining agnostic of the image size.

For the segmentation network, the three original labels are used, thus segmenting the region of oedema surrounding the tumour, the hyper-intense tumour regions, the necrotic core, and the background. Only the T1-weighted with gadolinium injection and the FLAIR sequences are used as input. For training, images are cropped around the ground truth tumour, all labels merged. The dimensions of the resulting images can vary independently along each axis from 32 to 96 voxels.

\subsection{Interface}

\begin{figure}[t]
    \centering
     \includegraphics[width=\textwidth]{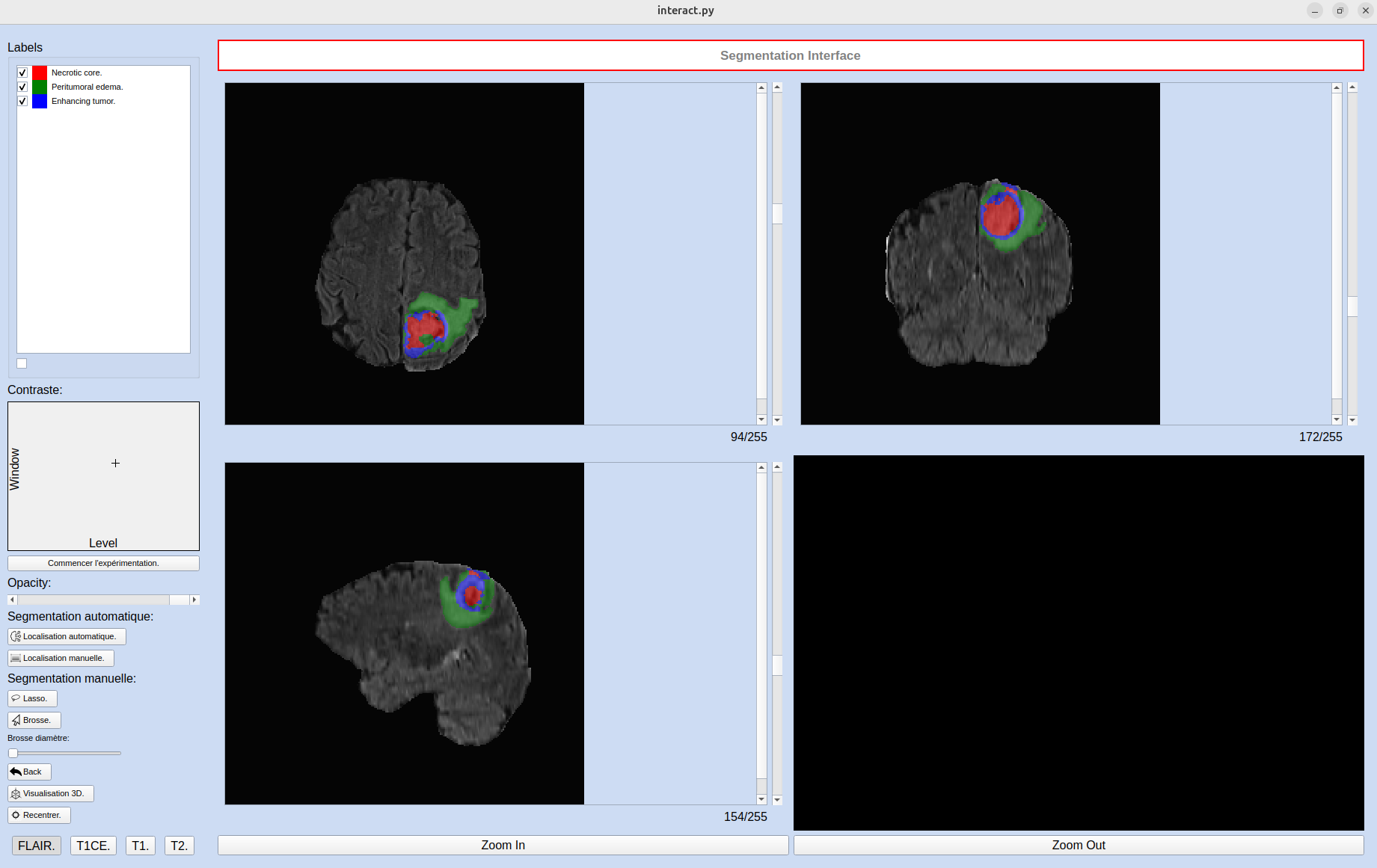}
    \caption{User-study interface}
    \label{fig:interface}
\end{figure}

Our segmentation interface allows users to segment gliomas using four different tools. One of them, the fully automatic segmentation, draws a bounding box automatically using the pre-segmentation network and performs the segmentation using the segmentation network. If needed, users can place their own bounding box with the bounding box tool, by drawing a rectangle on each view. Users can edit the segmentation using a variable size brush tool or a lasso tool.

\subsection{User experience experiment}

\subsubsection*{Participants}

The system was tested by two individuals: one neurosurgeon and one neuroradiologist, both male aged between 20 and 40 years. Both users have prior experience with commercial brain tumour segmentation interfaces, one twice a month and the other less frequently.

\subsubsection*{Exercises}

After following a tutorial, the participants were asked to complete the segmentation of all three labels on two different patient images, using any tool of their choice. The images are described as exercise 1 and 2.

\subsubsection*{Measures}

The total time to complete the experiment was measured along with the time spent using the brush tool for each time it is used, the number of voxels added or removed by label and the number of voxels removed using the brush tool, the number of uses of the lasso manual segmentation tool, the time to draw bounding boxes, the number of bounding boxes drawn, the number of uses of the fully automatic segmentation and the number of uses of the back function. The resulting segmentation as well as every step taken was saved.

\subsubsection*{Questionnaires}

Multiple questionnaires were completed by each participant.
One initial questionnaire concerned the participants' age, medical speciality, sub-speciality, professional status at the hospital, prior experience with brain tumour segmentation, for what kind of clinical practice and how much precision they need for this clinical practice.

At the end of each task, participants were asked for each label  whether they segmented it as exactly as possible, somewhat larger of somewhat smaller, if they felt they were more precise, as precise or less precise than usual and if they felt they were quicker, as quick or less quick than usual.

At the end of the experiment, the participants were given the Computer Usability Satisfaction Questionnaire (CSUQ) \cite{lewis} but not the part about the quality of provided informations, items 9-15 because the experiments were in person and the interface doesn't have error messages for now.

At the end of the experiment, users were asked their level of confidence in the two overall interaction mechanisms: the fully automatic tool and the manual localisation with automatic segmentation tool. Users were asked to choose between the following answers separately for each tool:
\begin{enumerate}
    \item Yes, really;
    \item Yes but I need more information to decide, for example about precision.;
    \item Yes but I need to experiment some more with the tool in order to decide;
    \item Largely no but if I had more information, for example about precision., it would help;
    \item Largely no but if I could experiment with the tool some more it would help;
    \item Not at all; and
    \item I have not used it.
\end{enumerate}
Users were allowed to provide an open comment explaining their decision, especially in cases 2 to 6 where additional functionalities, metrics, or testing could be suggested.

Users were also asked if this interface could be used in place of the one they currently used. The following answers to this question were available:
\begin{itemize}
    \item Yes;
    \item Yes for some functionalities but it lacks some other functionalities; and
    \item No.
\end{itemize}
Again, participants were also allowed to provide an open-form comment on their answer and encouraged to specific which interface they currently use for brain tumour segmentation in their clinical practice, if any.

\section{User Study Results}

\subsection*{Measures}
\begin{table}[h]
\centering
\begin{tabular}{|c|c|c|c|c|c|c|c|c|c|c|}
\hline
\makecell{Exer-\\cise} &
\makecell{Part-\\icipant} & 
\makecell{Brush\\(s)} & 
\makecell{Voxels\\added\\oedema\\/ necrotic \\ core \\/enhancing \\ tumour (nb)} & 
\makecell{Voxels\\removed (nb)} & 
\makecell{Lasso (nb)} & 
\makecell{Bounding \\ boxes\\(s)} & 
\makecell{Bounding \\ boxes\\(nb)} &
\makecell{Fully\\automatic\\(nb)} & 
\makecell{Back\\(nb)} & 
\makecell{Time\\(s)} \\
\hline 
\noalign{\vskip 1ex}
\hline
\multirow{2}{*}{1} & 1 & 22 & 0/0/0 & 22 & 0 & 0 & 0 & 1 & 0 & 322 \\ \cline{2-11}
& 2 & 0 & 0/0/0 & 0 & 0 & 0 & 0 & 1 & 0 & 97 \\
\hline
\noalign{\vskip 1ex}
\hline
\multirow{2}{*}{2} & 1 & 408 & 806/0/0 & 42 & 64 & 0 & 0 & 1 & 17 & 1392 \\ \cline{2-11}
& 2 & 0 & 0/0/0 & 0 & 0 & 0 & 0 & 1 & 0 & 112 \\
\hline
\end{tabular}
\label{tab:c}
\caption{Measures for the three segmentation exercises and three participants. Times are indicated in seconds (s) and numbers of voxels or uses as counts (nb).}
\end{table}

From these results, we can see that for both exercises, participant 2 performed the segmentation using only the fully automatic segmentation and spent 97 and 112 seconds overall, giving him some time to review it before clicking on the "I am satisfied of my segmentation" button. Participant two made a significant use of the brush and lasso tools, spending over 6 minutes using the brush tool making 806 edits to the oedema labels and using the lasso tool 64 times, indicating a missegmentation of the oedema label. The 1392s overall are not indicative, though, since the participant was interrupted during the exercise. The bounding box tool wasn't used by the participants, indicating a need for a clear error of localisation in future exercises.

\subsection*{Questionnaires}

For both the fully automatic segmentation tool and the manual localisation with automatic segmentation tool, which they used in the tutorial but not in the exercises, participant 1 said they trusted them but needed more experience with the interface to decide (Case 3). They said they think the interface could be used in place of the tool they currently used.
Participant 2 said they really trusted the fully automatic segmentation (Case 1) and hadn't used the bounding box segmentation (Case 7). They didn't use any other glioma segmentation tool in their clinical practice. We can see from these results that users have a good overall appreciation of the interface.

\begin{table}[h]
  \centering
  \begin{tabular}{|c|c|c|}
    \hline
    User & 1 & 2 \\ \hline
    System usefulness   & 5.875/7   & 6/7   \\ \hline
    Interface quality   & 5.667/7   & 5/7   \\ \hline
    Global & 6/7 & 6/7 \\ \hline 
  \end{tabular}
  \caption{C.S.U.Q. scores for the 2 participants.}
  \label{tab:s}
\end{table}

\subsection*{Suggestions from the participants}

Participants suggested that total tumour volume was more important than delineation of different labels in their clinical practice. They said that being able to perform a segmentation in a few seconds was important as they often need to do it during a consultation with the patient present.
They suggested adding the ability to adjust the contrast windowing for a selected region of interest. They also suggested a form of auto-completion of next slice, commonly referred to in the literature as interpolated segmentation \cite{baxter2018semiotics} in order to more efficiently propagate manual segmentation edits across slices. They also suggested the lasso tool should be able to remove voxels similarly to the functionality already present in the brush tool.
One participant also noted that their current practice does not require the segmentation of brain tumours into multiple compartments (i.e. hyper-intense tumour and necrotic core) which made the segmentation editing task more complicated than strictly needed for their use.

\section{Discussion and Limitations}

This work necessarily represents both algorithmic development, as can be understood as the development of a high-performance automated segmentation default mechanism, and human-computer interaction design, as can be understood through the observational user study performed in very ecologically-valid scenarios. This second focus is relatively uncommon in the literature regarding brain tumour segmentation but is nevertheless crucial for clinical trust and acceptability of machine learning based brain tumour segmentation algorithms.

The main limitation of our algorithmic framework for now is the poor specificity of the pre-segmentation network. Future work involving the use of data augmentation and the use of images containing tumours of all sizes might allow us to get rid of the false positive filtering. Data on the accuracy of our each segmentation step is much needed and will be provided in the interface once the segmentation networks are optimally trained.

The main limitation of our user study is definitely the small number of participants due to the difficulty in recruiting and scheduling user studies amongst what is a relatively small number of very busy individuals geographically spread out over multiple clinical centres. We are currently in the process of extending this recruitment and continuing the user study and thus more participants will be added to this experiment in the immediate future. 

Another limitation of our current user study is the lack of information regarding the efficiency of user interactivity, other measured by how close the segmentation gets to the final segmentation result after each individual use of an interaction mechanism \cite{wei2023towards}. Although our interface is currently designed to log said information, the fact that multiple interaction mechanisms are available and clinicians are free to chose what order they are used in, leads to a much more fundamentally heterogeneous notion of interaction mechanism efficiency that cannot be captured with single graphs which can be created for algorithms with only one interaction mechanism such as that presented by Wei \textit{et al.} \cite{wei2023towards}.

\section{Conclusion}
This interface represents, to the best of our knowledge, the first use-case of separable machine learning for larger-scale interactive segmentation correction for brain tumours as well as the first interactive brain tumour segmentation interface prototype to undergo expert user studies to understand which algorithmic elements are used and how they can be improved.

\bibliography{report} 

@inproceedings{giffard2025interactive,
  title={Interactive deep learning-based interface for separable subcortical anatomy segmentation.},
  author={Giffard, Enora and Jannin, Pierre and Baxter, John SH},
  booktitle={Medical Imaging 2025: Image-Guided Procedures, Robotic Interventions, and Modeling},
  volume={13408},
  pages={538--544},
  year={2025},
  organization={SPIE}
}

@article{baxter2018semiotics,
  title={The semiotics of medical image segmentation},
  author={Baxter, John SH and Gibson, Eli and Eagleson, Roy and Peters, Terry M},
  journal={Medical image analysis},
  volume={44},
  pages={54--71},
  year={2018},
  publisher={Elsevier}
}

@inproceedings{baxter2022combining,
  title={Combining simple interactivity and machine learning: a separable deep learning approach to subthalamic nucleus localization and segmentation in MRI for deep brain stimulation surgical planning},
  author={Baxter, John SH and Jannin, Pierre},
  journal={Journal of Medical Imaging},
  volume={9},
  number={4},
  pages={045001--045001},
  year={2022},
  publisher={Society of Photo-Optical Instrumentation Engineers}
}

@article{pellerino2022epidemiology,
  title={Epidemiology, risk factors, and prognostic factors of gliomas},
  author={Pellerino, Alessia and Caccese, Mario and Padovan, Marta and Cerretti, Giulia and Lombardi, Giuseppe},
  journal={Clinical and Translational Imaging},
  volume={10},
  number={5},
  pages={467--475},
  year={2022},
  publisher={Springer}
}

@article{wang2022virtual,
  title={Virtual reality-based education for patients undergoing radiation therapy},
  author={Wang, Liam J and Casto, Brian and Luh, Join Y and Wang, Samuel J},
  journal={Journal of Cancer Education},
  volume={37},
  number={3},
  pages={694--700},
  year={2022},
  publisher={Springer}
}

@inproceedings{hoebel2022know,
  title={Do I know this? segmentation uncertainty under domain shift},
  author={Hoebel, Katharina and Bridge, Christopher and Lemay, Andreanne and Chang, Ken and Patel, Jay and Rosen, Bruce and Kalpathy-Cramer, Jayashree},
  booktitle={Medical Imaging 2022: Image Processing},
  volume={12032},
  pages={261--276},
  year={2022},
  organization={SPIE}
}

@article{wei2023towards,
  title={Towards interactive deep-learning for tumour segmentation in head and neck cancer radiotherapy},
  author={Wei, Zixiang and Ren, Jintao and Korreman, Stine Sofia and Nijkamp, Jasper},
  journal={Physics and Imaging in Radiation Oncology},
  volume={25},
  pages={100408},
  year={2023},
  publisher={Elsevier}
}

@article{sailunaz2023brain,
  title={Brain tumor detection and segmentation: Interactive framework with a visual interface and feedback facility for dynamically improved accuracy and trust},
  author={Sailunaz, Kashfia and Bestepe, Deniz and Alhajj, Sleiman and {\"O}zyer, Tansel and Rokne, Jon and Alhajj, Reda},
  journal={Plos one},
  volume={18},
  number={4},
  pages={e0284418},
  year={2023},
  publisher={Public Library of Science San Francisco, CA USA}
}

@InProceedings{ronneberger2015unet,
author="Ronneberger, Olaf
and Fischer, Philipp
and Brox, Thomas",
editor="Navab, Nassir
and Hornegger, Joachim
and Wells, William M.
and Frangi, Alejandro F.",
title="U-Net: Convolutional Networks for Biomedical Image Segmentation",
booktitle="Medical Image Computing and Computer-Assisted Intervention -- MICCAI 2015",
year="2015",
publisher="Springer International Publishing",
address="Cham",
pages="234--241",
isbn="978-3-319-24574-4"
}

@misc{baid2021rsnaasnrmiccaibrats2021benchmark,
      title={The RSNA-ASNR-MICCAI BraTS 2021 Benchmark on Brain Tumor Segmentation and Radiogenomic Classification}, 
      author={Ujjwal Baid and Satyam Ghodasara and Suyash Mohan and Michel Bilello and Evan Calabrese and Errol Colak and Keyvan Farahani and Jayashree Kalpathy-Cramer and Felipe C. Kitamura and Sarthak Pati and Luciano M. Prevedello and Jeffrey D. Rudie and Chiharu Sako and Russell T. Shinohara and Timothy Bergquist and Rong Chai and James Eddy and Julia Elliott and Walter Reade and Thomas Schaffter and Thomas Yu and Jiaxin Zheng and Ahmed W. Moawad and Luiz Otavio Coelho and Olivia McDonnell and Elka Miller and Fanny E. Moron and Mark C. Oswood and Robert Y. Shih and Loizos Siakallis and Yulia Bronstein and James R. Mason and Anthony F. Miller and Gagandeep Choudhary and Aanchal Agarwal and Cristina H. Besada and Jamal J. Derakhshan and Mariana C. Diogo and Daniel D. Do-Dai and Luciano Farage and John L. Go and Mohiuddin Hadi and Virginia B. Hill and Michael Iv and David Joyner and Christie Lincoln and Eyal Lotan and Asako Miyakoshi and Mariana Sanchez-Montano and Jaya Nath and Xuan V. Nguyen and Manal Nicolas-Jilwan and Johanna Ortiz Jimenez and Kerem Ozturk and Bojan D. Petrovic and Chintan Shah and Lubdha M. Shah and Manas Sharma and Onur Simsek and Achint K. Singh and Salil Soman and Volodymyr Statsevych and Brent D. Weinberg and Robert J. Young and Ichiro Ikuta and Amit K. Agarwal and Sword C. Cambron and Richard Silbergleit and Alexandru Dusoi and Alida A. Postma and Laurent Letourneau-Guillon and Gloria J. Guzman Perez-Carrillo and Atin Saha and Neetu Soni and Greg Zaharchuk and Vahe M. Zohrabian and Yingming Chen and Milos M. Cekic and Akm Rahman and Juan E. Small and Varun Sethi and Christos Davatzikos and John Mongan and Christopher Hess and Soonmee Cha and Javier Villanueva-Meyer and John B. Freymann and Justin S. Kirby and Benedikt Wiestler and Priscila Crivellaro and Rivka R. Colen and Aikaterini Kotrotsou and Daniel Marcus and Mikhail Milchenko and Arash Nazeri and Hassan Fathallah-Shaykh and Roland Wiest and Andras Jakab and Marc-Andre Weber and Abhishek Mahajan and Bjoern Menze and Adam E. Flanders and Spyridon Bakas},
      year={2021},
      eprint={2107.02314},
      archivePrefix={arXiv},
      primaryClass={cs.CV},
      url={https://arxiv.org/abs/2107.02314}, 
}

@article{Menze2015BRATS,
  author    = {Bjoern H. Menze and Andras Jakab and Stefan Bauer and Jayashree Kalpathy-Cramer and Keyvan Farahani and Justin Kirby and Yuliya Burren and Nicole Porz and Johannes Slotboom and Roland Wiest and Levente Lanczi and Elizabeth Gerstner and Marc-Andr{\'e} Weber and Tal Arbel and Brian B. Avants and Nicholas Ayache and Patricia Buendia and D. Louis Collins and Nicolas Cordier and Jason J. Corso and Antonio Criminisi and Tilak Das and Herv{\'e} Delingette and {\c{C}}a{\u{g}}atay Demiralp and Christopher R. Durst and Michel Dojat and Senan Doyle and Joana Festa and Florence Forbes and Ezequiel Geremia and Ben Glocker and Polina Golland and Xiaotao Guo and Andac Hamamci and Khan M. Iftekharuddin and Raj Jena and Nigel M. John and Ender Konukoglu and Danial Lashkari and Jos{\'e} Anton{\'i}o Mariz and Raphael Meier and S{\'e}rgio Pereira and Doina Precup and Stephen J. Price and Tammy Riklin Raviv and Syed M. S. Reza and Michael Ryan and Duygu Sarikaya and Lawrence Schwartz and Hoo-Chang Shin and Jamie Shotton and Carlos A. Silva and Nuno Sousa and Nagesh K. Subbanna and Gabor Szekely and Thomas J. Taylor and Owen M. Thomas and Nicholas J. Tustison and Gozde Unal and Flor Vasseur and Max Wintermark and Dong Hye Ye and Liang Zhao and Binsheng Zhao and Darko Zikic and Marcel Prastawa and Mauricio Reyes and Koen Van Leemput},
  title     = {The Multimodal Brain Tumor Image Segmentation Benchmark (BRATS)},
  journal   = {IEEE Transactions on Medical Imaging},
  volume    = {34},
  number    = {10},
  pages     = {1993--2024},
  year      = {2015},
  doi       = {10.1109/TMI.2014.2377694},
  pmid      = {25494501},
  pmcid     = {PMC4833122},
  note      = {Epub 2014 Dec 4},
  publisher = {IEEE}
}

@article{Bakas2017TCGA,
  author    = {Bakas, Spyridon and Akbari, Hamed and Sotiras, Aristeidis and Bilello, Michel and Rozycki, Matthew and Kirby, Justin S. and Freymann, John B. and Farahani, Keyvan and Davatzikos, Christos},
  title     = {Advancing The Cancer Genome Atlas glioma MRI collections with expert segmentation labels and radiomic features},
  journal   = {Scientific Data},
  volume    = {4},
  pages     = {170117},
  year      = {2017},
  doi       = {10.1038/sdata.2017.117},
  pmid      = {28872634},
  publisher = {Nature Publishing Group}
}

@article{lewis,
author = {Lewis, James},
year = {2018},
month = {01},
pages = {1-9},
title = {Measuring Perceived Usability: The CSUQ, SUS, and UMUX},
volume = {34},
journal = {International Journal of Human-Computer Interaction},
doi = {10.1080/10447318.2017.1418805}
}
\bibliographystyle{spiebib} 

\end{document}